# An Advanced NLP Framework for Automated Medical Diagnosis with DeBERTa and Dynamic Contextual Positional Gating

Mohammad Ali Labbaf Khaniki[1], Sahabeh Saadati[2], Mohammad Manthouri[3]

[1]Faculty of Electrical Engineering, K.N. Toosi University of Technology, Tehran, Iran

[2]Department of Computer Engineering, North Tehran Branch, Islamic Azad University, Tehran, Iran

[3]Department Department of Electrical and Electronic Engineering, Shahed University, Tehran, Iran

mohamad95labafkh@gmail.com[1]

S.saadati@iau-tnb.ac.ir [2]

mmanthouri@shahed.ac.ir [3]

**Abstract:**

This paper presents a novel Natural Language Processing (NLP) framework for enhancing medical diagnosis through the integration of advanced techniques in data augmentation, feature extraction, and classification. The proposed approach employs back-translation to generate diverse paraphrased datasets, improving robustness and mitigating overfitting in classification tasks. Leveraging Decoding-enhanced BERT with Disentangled Attention (DeBERTa) with Dynamic Contextual Positional Gating (DCPG), the model captures fine-grained contextual and positional relationships, dynamically adjusting the influence of positional information based on semantic context to produce high-quality text embeddings. For classification, an Attention-Based Feedforward Neural Network (ABFNN) is utilized, effectively focusing on the most relevant features to improve decision-making accuracy. Applied to the classification of symptoms, clinical notes, and other medical texts, this architecture demonstrates its ability to address the complexities of medical data. The combination of data augmentation, contextual embedding generation, and advanced classification mechanisms offers a robust and accurate diagnostic tool, with potential applications in automated medical diagnosis and clinical decision support. This method demonstrates the effectiveness of the proposed NLP framework for medical diagnosis, achieving remarkable results with an accuracy of 99.78%, recall of 99.72%, precision of 99.79%, and an F1-score of 99.75%. These metrics not only underscore the model's robust performance in classifying medical texts with exceptional precision and reliability but also highlight its superiority over existing methods, making it a highly promising tool for automated diagnostic systems.

# 1. Introduction

Accurate and timely medical diagnosis is essential for effective healthcare, directly impacting patient outcomes and treatment quality. Delays or errors in diagnosis, especially for critical conditions like sepsis, cancer, or myocardial infarction, can worsen prognoses and lead to unnecessary treatments or fatalities. Traditional diagnostic methods remain labor-intensive, reliant on clinician expertise, and subject to variability due to subjective interpretations, cognitive biases, and human error [1]. Additionally, challenges such as time-consuming processes, resource limitations, and misinterpretations contribute to diagnostic inefficiencies. These issues are further exacerbated in regions with limited healthcare resources, highlighting the need for more reliable and efficient diagnostic solutions [2]. Recent advancements in Natural Language Processing (NLP) and deep learning have opened new possibilities for automating medical diagnosis, overcoming the limitations of traditional methods. AI-powered systems enhance scalability and accessibility, providing diagnostic support in underserved areas through mobile apps and telemedicine platforms. These models also help reduce diagnostic disparities by incorporating diverse data and serve as decision-support tools for clinicians, improving accuracy and reducing cognitive load without replacing human expertise [3].

The emergence of transformer architectures and attention mechanisms has revolutionized the field of NLP, offering significant improvements in model performance across a range of tasks. Transformers, introduced by Vaswani et al. (2017), leverage self-attention mechanisms to process input sequences in parallel rather than sequentially, enabling more efficient learning of long-range dependencies [4]. This parallelism not only accelerates training but also enhances the model's ability to capture complex relationships in data, overcoming the limitations of traditional sequential models such as Recurrent Neural Networks (RNNs) and Long Short-Term Memory

(LSTM) networks. The scalability of transformers also allows them to handle large datasets more effectively, making them suitable for a wide array of real-world applications, including large-scale language models and complex multi-task learning scenarios [5].

The attention mechanism, a core component of transformers, allows models to weigh the importance of different tokens in a sequence when making predictions. This capability enables more nuanced understanding and contextualization of language, as the model can focus on relevant parts of the input while disregarding less important information. Specifically, self-attention computes a weighted sum of all token representations in a given sequence, dynamically adjusting the focus on each token based on its relevance to other tokens. This mechanism is particularly effective in handling variable-length sequences, capturing intricate dependencies, and improving interpretability by allowing models to highlight which tokens influence their predictions [6]. The integration of transformers with attention mechanisms has led to substantial advancements in tasks such as machine translation, sentiment analysis, question answering, and text classification, making them the foundation for state-of-the-art NLP models such as BERT, GPT, and DeBERTa [7], [8].

Data augmentation plays a crucial role in sentiment analysis and text classification by enhancing the diversity and volume of training data, which helps improve model generalization and performance. In these tasks, the availability of labeled data is often limited, and augmenting the dataset through techniques such as paraphrasing, back-translation, and noise injection allows models to learn from a broader range of linguistic variations [9]. By simulating different ways in which sentiments or categories can be expressed, data augmentation mitigates the risk of overfitting and enables the model to become more robust in handling unseen or rare expressions. This process is particularly important for dealing with imbalanced datasets, where it can help prevent bias toward the majority class and ensure that the model is equally sensitive to all categories or sentiments. Ultimately, data augmentation is essential for building high-performing text classification models, especially in real-world applications where data diversity is critical. In medical diagnosis, data augmentation techniques such as back-translation and paraphrasing are essential for increasing the diversity of clinical text data, enabling models to better generalize across varied symptom descriptions and improving diagnostic accuracy [10].

This paper introduces a novel approach for medical diagnosis using a combination of advanced techniques in data augmentation, feature extraction, and classification. The methodology can be summarized as follows:

1. **Back-Translation for Data Augmentation**:

    The model utilizes a back-translation technique to augment data diversity. This process involves translating text from English to German and subsequently re-translating it back to English, thereby generating paraphrased versions of the original data. This approach aids in mitigating overfitting and enhances the robustness of the classifier.

2. **DeBERTa with Dynamic Contextual Positional Gating (DCPG):**

    The DeBERTa model serves as the backbone for feature extraction, leveraging its disentangled attention mechanism and enhanced decoding to capture fine-grained contextual and positional relationships in the input text. This allows the model to generate high-quality embeddings that are crucial for downstream classification tasks. Building on DeBERTa's foundation, the extended version, DeBERTa with DCPG, introduces a novel gating mechanism that adaptively adjusts the interaction between content and positional information based on the semantic context. Unlike the original DeBERTa, which relies on

fixed interactions to separate content and position, DCPG dynamically modulates the positional influence in a context-aware manner. This enhancement improves the model's flexibility and allows the attention mechanism to focus more effectively on relevant features. By combining relative positional embeddings with dynamic gating, DCPG achieves a balanced integration of positional and semantic information, making it highly effective for tasks that demand a detailed understanding of both content and structure.

3. **Attention-Based Feedforward Neural Network (ABFNN) for Classification**:

   The ABFNN serves as the classifier within the proposed pipeline. This innovative architecture integrates attention mechanisms with a feedforward neural network, enabling the model to selectively focus on the most pertinent features of the input data. By effectively leveraging the hierarchical representations provided by DeBERTa, the ABFNN enhances the overall classification accuracy.

This architecture is applied to medical diagnosis by classifying symptoms, clinical notes, and related data into predefined categories or conditions. The back-translation augmentation enhances data resilience, while DeBERTa generates robust text representations. The ABFNN component ensures precise decision-making. Collectively, these elements address the complexities inherent in medical text data, providing a robust and accurate diagnostic tool. Comparative results demonstrate the superiority of the proposed method over Medical Concept Normalization—Bidirectional Encoder Representations from Transformers (MCN-BERT) [11].

## 2. Background and Related Work

This section provides an overview of the medical diagnosis dataset, detailing its structure, features, and relevance to the task of disease classification. Additionally, a comprehensive literature review is presented, summarizing key studies and methodologies in the field of medical diagnosis using NLP techniques.

## 2.1. Medical Diagnosis Dataset Overview

The Symptom2disease dataset is a comprehensive collection of symptom-disease associations designed for medical text analysis and disease prediction tasks. It comprises 1200 datapoints, each containing two key elements: a disease label and a corresponding natural language description of symptoms. The dataset covers 24 distinct diseases, with each disease represented by 50 unique symptom descriptions. This balanced distribution results in a total of 1200 entries. The diseases included in the dataset span a wide range of medical conditions, including dermatological disorders (e.g., Psoriasis, Acne), infectious diseases (e.g., Typhoid, Chicken pox, Dengue), respiratory conditions (e.g., Common Cold, Pneumonia, Bronchial Asthma), and chronic illnesses (e.g., Hypertension, Diabetes) [11].

- label: This column contains the disease labels, representing the 24 different medical conditions.
- text: This column includes natural language descriptions of symptoms associated with each disease label.

The Symptom2disease dataset is publicly available on Kaggle in https://www.kaggle.com/datasets/niyarrbarman/symptom2disease/, making it accessible to researchers and data scientists interested in exploring the relationship between symptoms and diseases in a structured format.

## 2.2. Literature review

Machine learning has found widespread applications across various domains, including healthcare, finance, and autonomous systems, where it enables data-driven decision-making and automation. Esmaeili et al.'s study examines recent advancements in AI-driven customer service technologies, focusing on chatbots and virtual assistants that provide personalized support, analyze consumer data for predictive insights, and enhance overall customer experiences [12]. Li et al. developed and evaluated machine learning models to predict hospital mortality in mechanically ventilated ICU patients, using 32 selected features from the MIMIC-III database, with the CatBoost model achieving the highest performance [13]. Sepanloo et al. propose a multi-sensor augmented reality system for nursing education that integrates various sensors with AR technology to provide precise analysis and feedback during training simulations, enhancing the learning experience for nursing students [14]. Khodayari Gharanchaei's study compares various machine learning methods for predicting credit card defaults using a dataset of 30,000 Taiwanese clients, evaluating different approaches to classify defaulters and non-defaulters while considering the impact of including or excluding borderline cases [15]. This study by Mohammadjafari et al. explores using eye tracking to detect cognitive load in complex virtual reality training environments, employing machine learning models to predict cognitive load from eye movement data and demonstrating the potential for developing adaptive VR training systems [16]. This study proposes a novel approach for safe robot navigation in unmapped environments using composite control barrier functions, addressing both state and input constraints while leveraging real-time perception feedback to construct local CBFs, enabling optimal control generation for obstacle avoidance and constraint satisfaction [17].

Deep learning has revolutionized numerous fields, such as image recognition, natural language processing, and speech synthesis, by enabling highly accurate models through neural networks with multiple layers. Shafee et al. evaluate the performance of various LLM chatbots in OSINT-based cybersecurity tasks, finding that while some models like GPT-4 show promise in binary classification, all tested chatbots have limitations in named entity recognition compared to specialized models [18]. Benchari and Totaro's study presents an integrated U-Net and ResNet50 architecture for MRI brain cancer image detection, combining the strengths of both models to improve segmentation and classification performance in identifying brain tumors from MRI scans [19]. Mohammadjafari et al. provide a comprehensive review of LLM-based text-to-SQL systems, tracing their evolution from early rule-based models to advanced approaches using large language models and retrieval augmented generation (RAG), while examining key aspects like benchmarks, evaluation methods, and challenges in the field [20]. Merikhipour et al. propose a novel approach for transportation mode detection that combines spatial attention-based transductive LSTM with off-policy feature selection, enhancing the model's ability to capture spatial-temporal dependencies and select relevant features from GPS trajectories [21].

NLP and LLMs are revolutionizing healthcare by enhancing various aspects of patient care, clinical workflows, and medical research. These AI technologies are being applied to analyze vast amounts of medical literature, clinical records, and scientific papers, providing valuable insights to aid in accurate diagnoses, treatment planning, and informed decision-making. This study compares the performance of WE-LSTM networks and a WizardLM-powered chatbot called DiabeTalk for diagnosing diabetes from medical texts, likely evaluating their accuracy, efficiency, and potential applications in clinical settings [22]. This study proposes a Graph Attention Network (GAT) model combined with curriculum learning for classifying depression from online media, utilizing graph-based techniques to handle increasingly challenging training samples and improve mental healthcare classification performance [23]. This study proposes a medical specialty prediction model using a domain-specific pre-trained BERT to analyze patient-side medical question texts. The researchers fine-tuned the model to predict medical specialty labels from a dataset of medical questions, demonstrating improved performance compared to competitive models and showcasing its potential benefits for hospital patient management and specialty recommend [24]. This study explores using the BERT language model for multi-criteria classification of scientific articles, specifically employing SciBERT to classify MEDLINE abstracts based on multiple selection criteria. The researchers compared different ensemble architectures, including a novel cascade ensemble, to a single integrated model, finding that the cascade ensemble achieved higher precision and F-measure, while the single model performed better for tasks requiring high fixed recall like systematic reviews [25]. This study proposes a role-distinguishing BERT model for medical dialogue systems in smart cities, aiming to improve the accuracy of intent recognition by differentiating between patient and doctor roles in conversations [26]. This study proposes a role-distinguishing BERT model for medical dialogue systems in smart cities, aiming to improve the accuracy of intent recognition by differentiating between patient and doctor roles in conversations, contributing to the development of sustainable smart city healthcare services [27]. This study explores the application of BERT-based models for biomedical natural language inference on clinical trials, likely aiming to improve the accuracy and efficiency of processing and understanding clinical trial reports and related medical texts [28]. This study proposes DeBERTa-BiLSTM, a deep learning model combining DeBERTa and BiLSTM networks for multi-label classification of Arabic COVID-19 questions. The model demonstrates promising performance in categorizing medical questions into multiple categories, achieving high accuracy and precision, which could significantly benefit telehealth services and automated medical question-answering systems [29]. This study compares the performance of RoBERTa,

CNN, and ChatGPT models in automatically detecting unexpected findings in radiology reports, demonstrating that a fine-tuned RoBERTa model outperforms both CNN and ChatGPT for this specific task [30]. This study proposes a novel cross-attention approach called PSAT that incorporates clinical practice guidelines for depression into transformer models, aiming to enhance explainability and generate clinician-understandable explanations for depression diagnosis from unstructured text data [31]. This study proposes a novel automatic medical report generation model that combines a Detector Attention Module to fuse visual and location features with a GPT-based Word LSTM for improved report generation from medical images [32].

## 3. Methodology

In this section, we outline the methodologies employed in the proposed framework for enhancing medical diagnosis using NLP. The approach integrates three key techniques: back-translation for data augmentation, DeBERTa with DCPG for feature extraction, and an ABFNN for classification. Each of these components plays a crucial role in improving the model's accuracy, robustness, and contextual understanding in the medical domain.

### 3.1. Back-Translation for Data Augmentation

To enhance the generalizability and robustness of the proposed model, a back-translation technique was employed for data augmentation. This approach involves translating input text into an intermediate language and subsequently translating it back into the original language. By rephrasing the text through the translation pipeline, back-translation introduces lexical diversity into the dataset, enriching it with variations in word choice and sentence structure. Importantly, this technique ensures semantic preservation, maintaining the original meaning of the text—a critical factor for tasks such as medical diagnosis. The added variability in the dataset reduces the model's reliance on specific patterns in the training data, thereby mitigating overfitting and improving its ability to generalize effectively to unseen data.

The back-translation process begins with an initial translation of the input text from English into German using the MarianMT model (Helsinki-NLP/opus-mt-en-de) [33]. This neural machine translation system, built on the Transformer architecture, has been fine-tuned on extensive parallel corpora from the OPUS project. By leveraging diverse open-domain texts, the model achieves high-quality translations suited for general-purpose applications. Subsequently, the German-translated text undergoes reverse translation back into English using the complementary MarianMT model (Helsinki-NLP/opus-mt-de-en). This second translation pass ensures further diversification of the text while preserving its semantic integrity, enriching the dataset with varied lexical and syntactic structures [34].

### 3.2. DeBERTa with DCPG

Disentangled attention in BERT refers to separating content and positional information within the attention mechanism to improve how the model understands relationships between tokens. Instead of combining token embeddings with positional embeddings directly, as in standard transformers, disentangled attention treats semantic content and relative positional information independently, enabling the model to learn distinct interactions for content-to-content, content-to-position, and position-to-content relationships. This separation enhances the model's flexibility in handling tasks where positional dependencies vary in importance, such as long-sequence understanding or tasks with complex contextual relationships. By explicitly modeling these interactions, disentangled attention provides a more nuanced representation of token relationships, improving performance on a variety of NLP tasks.

### 3.2.1. Input Representation

Let an input sequence contain n tokens. The model utilizes two embedding matrices to represent semantic content and relative positional relationships:

- **Content Embeddings**:

$$C = [c_1, c_2, \dots, c_n]^T \in \mathbb{R}^{(n \times d)} \tag{1}$$

where $c_i \in \mathbb{R}^{(d)}$ represents the semantic content of the $i$-th token. The content embeddings matrix $C$ encodes the semantic meaning of tokens in the input sequence. These embeddings are typically obtained from a token embedding layer that maps vocabulary tokens into $d$ dimensional dense vectors

- **Relative Positional Embeddings**:

$$P = [p_1, p_2, \dots, p_n]^T \in \mathbb{R}^{(k_{max} \times d)} \tag{2}$$

where $p_{|i-j|} \in \mathbb{R}^{(d)}$ encodes the clipped relative distance $|i-j|$ between tokens $i$ and $j$, with $k_{max}$ denoting the maximum allowed distance. To ensure manageable computational complexity and memory usage, the relative distance $|i-j|$ is clipped to a maximum allowable distance $k_{max}$, meaning all distances greater than $k_{max}$ are treated as $k_{max}$. This clipping mechanism balances model expressiveness with computational efficiency while enabling the model to capture meaningful relative positional relationships.

### 3.2.2. Query and Key Projections

The model learns separate projection matrices to compute queries (**Q**) and keys (**K**):

- **Content-Specific Projections**:

The content embeddings matrix $C$ is used to compute content-based queries ($\mathbf{Q}_C$) and keys ($\mathbf{K}_C$) through linear projections:

$$\mathbf{Q}_C = CW_Q^C, \quad \mathbf{K}_C = CW_K^C \tag{3}$$

where $W_Q^C, W_K^C \in \mathbb{R}^{(d \times d)}$ are learnable weight matrices of shape. These projections enable the model to extract semantic relationships between tokens in the sequence.

- **Position-Specific Projections**:

The same content embeddings matrix $C$ is also used to compute position-based queries ($\mathbf{Q}_P$) and keys ($\mathbf{K}_P$) through separate projection matrices:

$$\mathbf{Q}_P = CW_Q^P, \quad \mathbf{K}_P = CW_K^P \tag{4}$$

where $W_Q^P, W_K^P \in \mathbb{R}^{(d \times d)}$ are independent trainable weight matrices. These projections allow the model to focus on relative positional relationships between tokens, leveraging the positional embedding information indirectly.

### 3.2.3. Disentangled Attention Components

In disentangled attention mechanisms, the attention score for tokens $i$ and $j$ is decomposed into three distinct components, each capturing a specific interaction type between the content and positional information of the tokens:

- **Content-to-Content Interaction** $A_{i,j}^{(cc)}$:

This component represents the semantic relationships between the content of tokens $i$ and $j$. The attention score is computed as:

$$A_{i,j}^{(cc)} = \frac{\boldsymbol{Q}_{C_i} K_{C_j}^T}{\sqrt{d}} \tag{5}$$

Here, $\boldsymbol{Q}_{C_i}$ and $\boldsymbol{K}_{C_j}$ are the content-based query and key projections for tokens $i$ and $j$, respectively. This term measures the alignment between their semantic content. capturing semantic relationships between tokens.

- **Content-to-Position Interaction $A_{i,j}^{(cp)}$:**

This component models how the semantic content of token iii relates to the positional embedding of token $j$. The score is given by:

$$A_{i,j}^{(cp)} = \frac{\boldsymbol{Q}_{C_i} K_{P_{|i-j|}}^T}{\sqrt{d}} \tag{6}$$

Here, $\boldsymbol{Q}_{C_i}$ is the content-based query for token i, and $\boldsymbol{K}_{P_{|i-j|}}$ is the key projection for the relative positional embedding $|i-j|$. This term captures how a token's semantic content interacts with the positional context of another token.

- **Position-to-Content Interaction $A_{i,j}^{(pc)}$:**

This component models how the positional embedding of token $i$ interacts with the semantic content of token $j$. The attention score is computed as:

$$A_{i,j}^{(pc)} = \frac{\boldsymbol{Q}_{P_{|i-j|}} K_{C_j}^T}{\sqrt{d}} \tag{7}$$

Here, $\boldsymbol{Q}_{P_{|i-j|}}$ is the query projection for the relative positional embedding $|i-j|$, and $\boldsymbol{K}_{C_j}$ is the content-based key for token $j$. This term captures how positional information influences the understanding of semantic content. These three components collectively disentangle the roles of semantic content and positional information in the attention mechanism. By explicitly separating content and position interactions, the model gains finer control over how it interprets token relationships, leading to better performance and interpretability in tasks where both semantic and positional relationships are crucial.

### 3.2.4. Dynamic Contextual Gating Mechanism

To dynamically modulate positional influence based on contextual relevance, we introduce a learnable gating scalar.

- **Gating Scaler**:

The gating scalar $\alpha_{i,j}$ is a learnable value that modulates the influence of positional relationships between tokens $i$ and $j$ based on their semantic context. It is computed as:

$$\alpha_{i,j} = \sigma\left(Q_{C_i} W_g K^T_{P_{|i-j|}}\right) \tag{8}$$

Where $W_g \in \mathbb{R}^{(d \times d)}$ is trainable matrix, and $\sigma(.)$ is the sigmoid function, ensuring the gating scalar $\alpha_{i,j} \in [0,1]$. This formula dynamically adjusts the influence of positional relationships based on the semantic context, making the model more flexible in capturing complex token dependencies.

- **Gated Content-to-Position Interaction**:

The Gated Content-to-Position Interaction is the process where the content-to-position interaction matrix $A^{(cp)}_{i,j}$ is modulated by the gating scalar $\alpha_{i,j}$. It is defined as:

$$\tilde{A}^{(cp)}_{i,j} = \alpha_{i,j} \cdot A^{(cp)}_{i,j} \tag{9}$$

This operation allows the model to adaptively adjust the strength of token interactions based on their contextual meaning and positional distance, making the model more flexible and context-aware.

### 3.2.5. Total Attention Score and Output

The final attention score combines the three components with gradient-stabilizing scaling:

$$A_{i,j} = \frac{A_{i,j}^{(cc)} + \tilde{A}_{i,j}^{(cp)} + A_{i,j}^{(pc)}}{\sqrt{3d}} \tag{10}$$

To convert the raw attention scores into probability-like values, softmax normalization is applied across all tokens for each $i$. This step ensures that the sum of the attention weights for each token equals 1, making them interpretable as probabilities

$$\text{AttentionWeights}_{i,j} = \text{softmax}(A_{i,j}) \tag{11}$$

The softmax function transforms the attention scores into normalized attention weights that highlight the relative importance of each token when computing the final output. The attention output for token $i$ is calculated as the weighted sum of value projections for all tokens:

$$\text{Output}_i = \sum_{j=1}^{n} \text{AttentionWeights}_{i,j} \cdot V_j \tag{12}$$

Where $V_j = C_j W_V$ represents the value projection of token $j$, and $\boldsymbol{W}_V \in \mathbb{R}^{(d \times d)}$ is a learnable weight matrix.

DeBERTa's with DCPG significantly enhances the capabilities of DeBERTa's disentangled attention by introducing a dynamic mechanism that adaptively modulates the influence of positional interactions based on semantic context. Unlike DeBERTa, which statically weights content-to-position (C→P) and position-to-content (P→C) interactions uniformly across all token pairs, DCPG employs a learnable gating scalar $\alpha_{i,j}$, conditioned on the semantic embeddings of tokens $i$ and $j$. This gate dynamically suppresses or amplifies positional dependencies when semantic context dominates (e.g., resolving lexical ambiguities like "bank" near "river") or when syntactic structure is critical (e.g., clause boundaries), respectively. By integrating this context-aware flexibility, DCPG achieves finer-grained control over disentangled interactions, reducing over-reliance on positional biases and improving performance on tasks requiring nuanced semantic reasoning, such as coreference resolution or topic modeling. The addition of a lightweight gate matrix $\boldsymbol{W}_g$ ensures computational efficiency while enabling the model to outperform DeBERTa in adaptability and accuracy across diverse linguistic scenarios. The Comprehensive comparison between the BERT, DeBERTa, and the proposed DeBERTa with DCPG models is given in Table 1.

**Table 1.** Comprehensive comparison between the BERT, DeBERTa, and the proposed DeBERTa with DCPG models.

| Models \ Aspect | BERT | DeBERTa | DeBERTa with DCPG |
|---|---|---|---|
| **Interaction Terms** | Single entangled term: $A_{i,j}$ = Content + Position | Three fixed terms: 1. C→C  2. C→P  3. P→C. | Retains DeBERTa's three terms but **adaptively reweights** C→P via a learned gate $\alpha_{i,j}$. |
| **Positional Encoding** | Absolute positional embeddings. | Relative positional embeddings (distance-based). | Relative positional embeddings + dynamic gating. |
| **Gating Mechanism** | None. | None. | Introduces $\alpha_{i,j} = \sigma\left(\boldsymbol{Q}_{c_i}\boldsymbol{W}_g K^T_{P_{|i-j|}}\right)$. |
| **Attention Score** | $A_{i,j} = \frac{(c_i + p_i)\boldsymbol{W}_Q(c_i + p_i)W_k^T}{\sqrt{d}}$ | $A_{i,j} = \frac{A^{(cc)}_{i,j} + A^{(cp)}_{i,j} + A^{(pc)}_{i,j}}{\sqrt{3d}}$ | $A_{i,j} = \frac{A^{(cc)}_{i,j} + \tilde{A}^{(cp)}_{i,j} + A^{(pc)}_{i,j}}{\sqrt{3d}}$ |
| **Key Interaction** | Entangles content and position. | Disentangles content/position into three additive terms. | Disentangles terms + modulates C→P via $\alpha_{i,j}$. |
| **Gate Computation** | None. | None. | $\alpha_{i,j} = \sigma\left(\boldsymbol{Q}_{c_i}\boldsymbol{W}_g K^T_{P_{|i-j|}}\right)$ |

| | | | |
|---|---|---|---|
| Positional Adaptation | Static: Fixed absolute positions. | Static: Relative positions but fixed interaction weights. | Dynamic: Adjusts positional influence via semantic context. |
| Flexibility | Limited: Position and content are fused. | Improved: Explicit disentanglement of content/position. | Context-aware: Suppresses/enhances positional bias based on semantics. |

### 3.3. ABFNN

The ABFNN is a novel neural architecture designed to enhance feature integration and classification performance through the combination of hierarchical feature extraction, multi-head self-attention, and multi-scale integration mechanisms. The ABFNN incorporates multiple processing branches to extract diverse representations from input features, followed by a multi-head self-attention mechanism that dynamically identifies and emphasizes important features. To further refine the representations, the architecture integrates residual connections, promoting efficient learning and mitigating gradient vanishing issues in deeper layers. A multi-scale feature integration module combines attention-enhanced and hierarchical features to capture both global and local information effectively. By leveraging attention mechanisms and hierarchical processing, ABFNN achieves improved performance and interpretability compared to traditional feedforward networks. The key components of ABFNN are:

1. Hierarchical Feature Extraction: Two branches extract diverse features:

$$h_{branch_1} = \text{ReLU}(W_1 x + b_1), \quad h_{branch_2} = \text{ReLU}(W_2 x + b_2) \tag{13}$$

$$h_{combined} = h_{branch_2} + h_{branch_1} \tag{14}$$

2. Self-Attention Mechanism: Focuses on important features:

$$\text{Attention}(Q, K, V) = \text{Softmax}\left(\frac{QK^T}{\sqrt{d_k}}\right) \times V. \tag{15}$$

with $Q, K, V = W_Q h_{combined}, W_K h_{combined}, W_V h_{combined}$

3. Residual Connection and Multi-Scale Integration:

$$h_{residual} = \text{ReLU}(W_r \text{Attention} + b_r) + \text{Attention} \tag{16}$$

$$h_{multi-scale} = \text{ReLU}(W_m [h_{residual}, h_{combined}] + b_m) \tag{17}$$

4. Output Layer:

$$y = \text{softmax}(W_{out} h_{multi-scale} + b_{out}) \tag{18}$$

This design enhances feature representation and classification, making it suitable for complex datasets, such as outputs from models like BERT. The Figure 1 illustrates the architecture of the ABFNN, highlighting its integrated attention mechanism designed to focus on the most relevant features for improved classification accuracy in medical diagnosis.

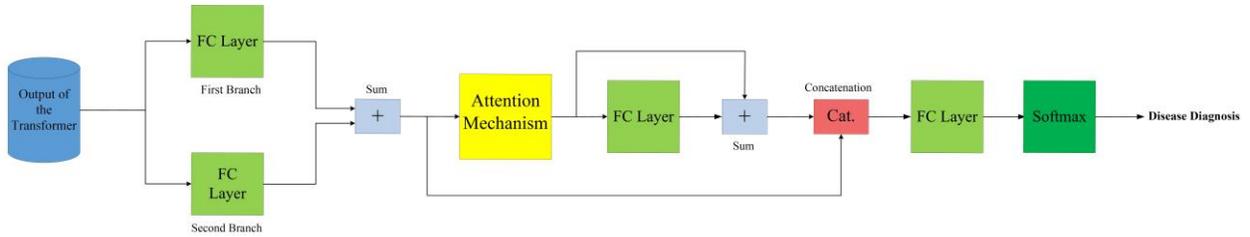

Figure 1: ABFNN Architecture.

## 3.4. The proposed Medical Diagnosis Method

This subsection provides an overview of the proposed method, which consists of five key stages:

1. **Symptom Descriptions**: Patients' symptoms are provided in textual form, serving as the primary input for the diagnosis model.

2. **Back-Translation for Data Augmentation**: To enhance the diversity and robustness of the dataset, the input texts are translated to German and back to English using the MarianMT model. This introduces lexical variety while preserving semantic meaning, improving the model's ability to generalize.

3. **Data Collection and Preprocessing**: The collected symptom descriptions are cleaned, tokenized, and prepared for input into the model, ensuring high-quality and standardized data for training.

4. **DeBERTa with DCPG**: The input data is processed using the DeBERTa model, which employs disentangled attention and positional gating to capture contextual relationships and subtle nuances in the text for accurate feature representation.

5. **ABFNN**: The output embeddings from DeBERTa are fed into the ABFNN, which uses an integrated attention mechanism to focus on the most relevant features and improve classification accuracy for medical diagnosis.

The schematic representation of the proposed method is illustrated in Figure 2.

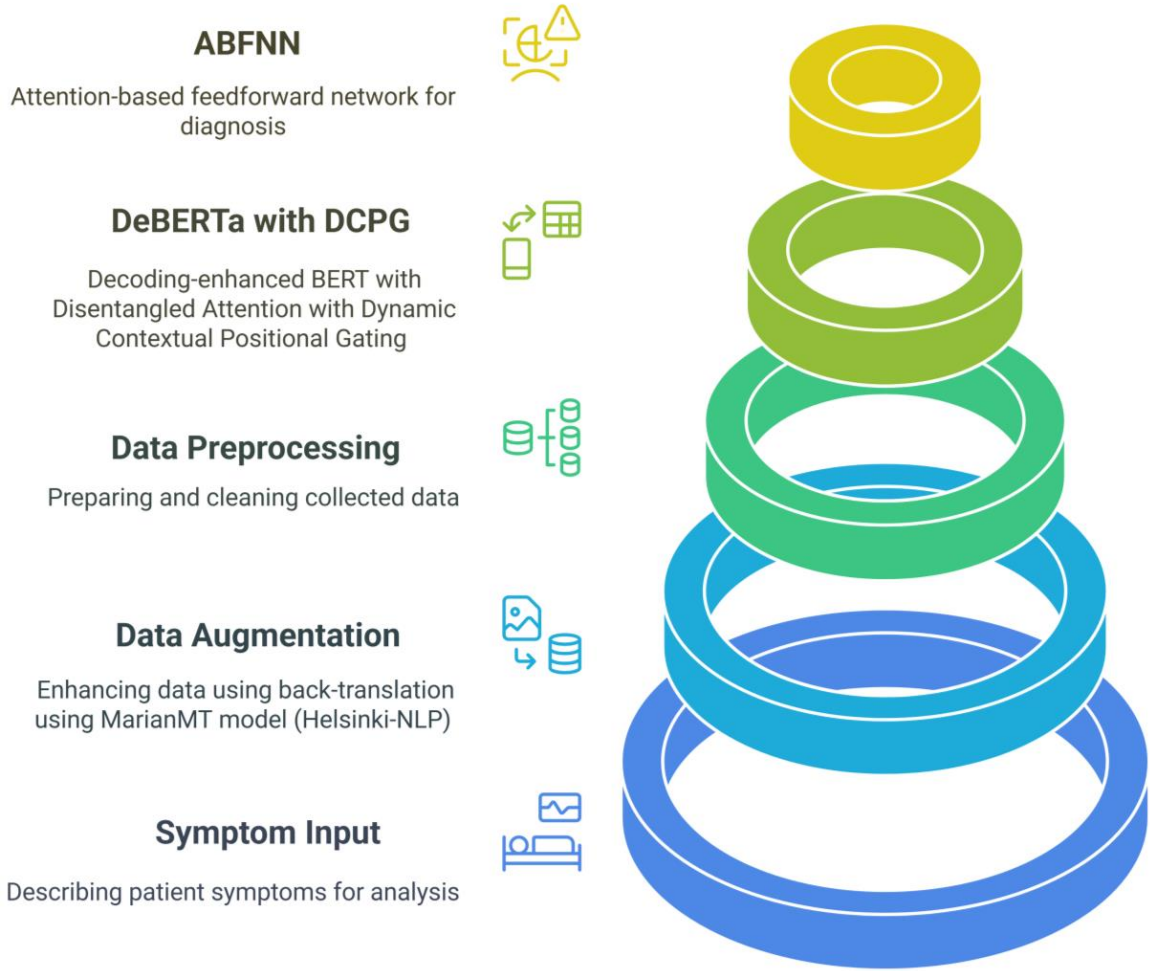

Figure 2: Schematic of the medical diagnosis network phases.

## 4. Simulations

This section presents a comprehensive evaluation of the performance of several deep learning models, including the BERT, DeBERTa, and the methods which is presented in [11]. we utilized the model with the following hyper parameters:

- Number of Layers: 12 layers in the transformer encoder.

- Hidden Size: The hidden size of each layer was set to 768, determining the dimensionality of the token embeddings.
- Number of Attention Heads: The model employed 12 attention heads in each self-attention layer, allowing it to focus on different aspects of the input sequence simultaneously.
- Feedforward Size: The intermediate size in the feedforward layers was set to 1024, providing the network with sufficient capacity to learn complex transformations of the hidden states.

Additionally, all the feedforward layers in the model employed the ReLU activation function, which introduces non-linearity and enables the network to capture more complex patterns in the data. This configuration was chosen to strike a balance between computational efficiency and model performance, leveraging the strengths of the transformer architecture while ensuring sufficient expressiveness for the task at hand.

## 4.1. Accuracy

This subsection utilizes accuracy as a key evaluation metric to measure the performance of the models. The formula for accuracy is expressed as follows:

$$Accuracy = \frac{TP + TN}{TP + TN + FP + FN} \tag{17}$$

In the context of binary classification, True Positives (TP) represent the count of positive instances that are correctly identified as positive, while True Negatives (TN) denote the count of negative instances accurately classified as negative. Conversely, False Positives (FP) refer to negative instances that are mistakenly predicted as positive, and False Negatives (FN) indicate positive instances that are incorrectly identified as negative.

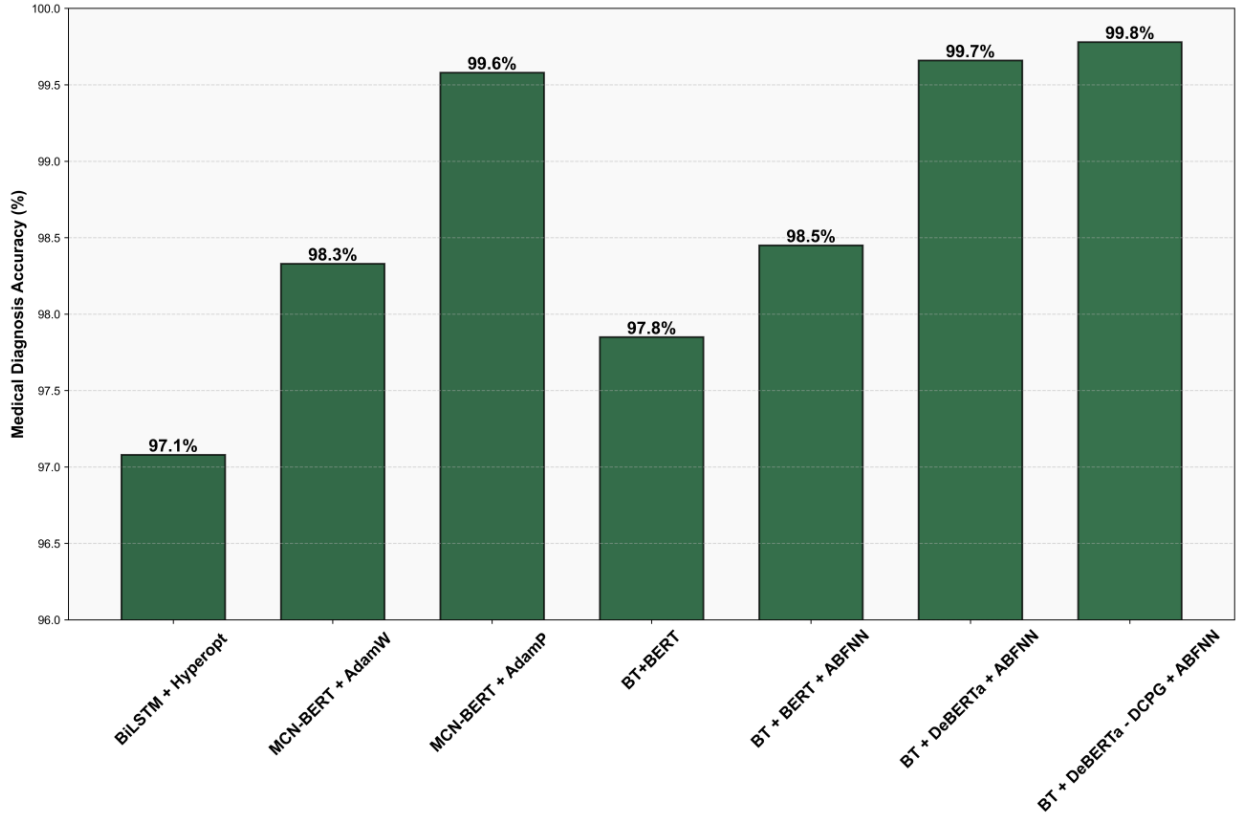

Figure 3: Accuracy Comparison of Medical Diagnosis Methods Between the Proposed Model and Existing Studies [11].

Figure 3 presents a bar graph comparing the accuracy of various deep learning models in medical diagnosis. The combination of BT + DeBERTa + DCPG + ABFNN achieves the highest accuracy of 99.8%, followed by BT + DeBERTa + ABFNN at 99.7%. The lowest-performing model, BiLSTM + Hyperopt, records an accuracy of 97.1%. Models incorporating DeBERTa and ABFNN demonstrate superior performance, emphasizing their effectiveness in medical text analysis.

This high accuracy stems from the synergy between advanced techniques addressing different challenges in medical diagnosis. Back-Translation enhances data diversity by paraphrasing symptom descriptions, improving model generalization and reducing overfitting. DeBERTa, combined with DCPG, disentangles content and positional information, dynamically adjusting their influence to better interpret complex symptom contexts. Meanwhile, ABFNN employs attention mechanisms to prioritize diagnostically significant features, ensuring accurate predictions. Together, these components form a robust framework that handles variability, context, and feature importance, achieving better accuracy compared to MCN-BERT and other models [11].

## 4.2. Confusion Matrix

The confusion matrix in Figure 4 illustrates the classification performance of the proposed DeBERTa with DCPG model on a medical diagnosis task.

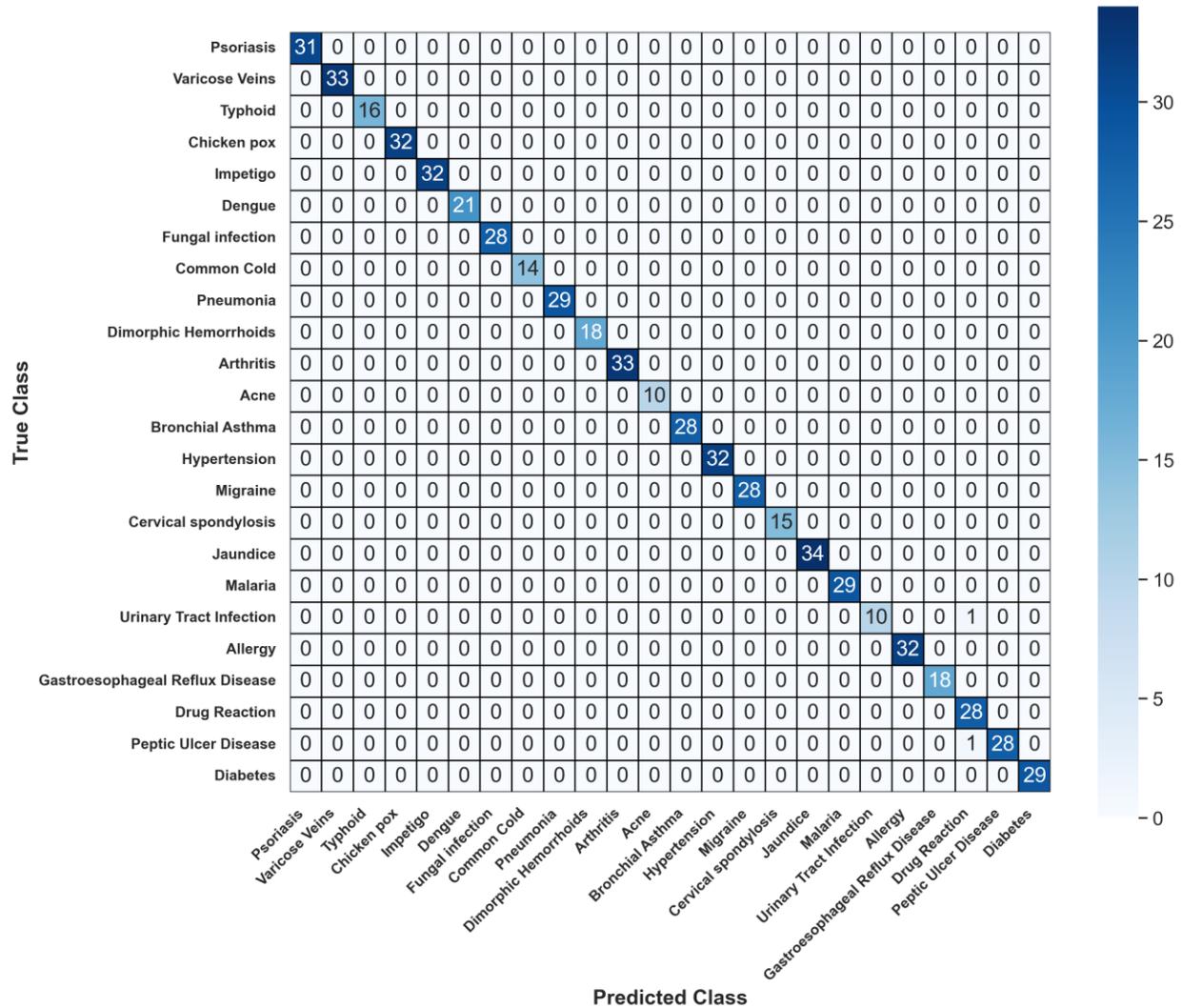

Figure 4: Confusion matrix using the proposed DeBERTa with DCPG.

The confusion matrix in Figure 4, provides an insightful evaluation of the DeBERTa with DCPG model's classification performance across 24 medical conditions, comparing predicted disease classes against true labels. The matrix's structure, characterized by diagonal dominance, highlights the model's impressive ability to correctly classify conditions such as Psoriasis, Diabetes, and Malaria. Misclassifications, evident in off-diagonal cells, are more prominent in cases of overlapping symptoms, such as Common Cold vs. Pneumonia, where respiratory similarities pose challenges, or Jaundice vs. Drug Reaction, where shared symptoms like yellowing skin lead to errors. Despite these, the high counts along the diagonal underscore the model's robustness and accuracy for most conditions.

Key strengths of the model include its dynamic positional adaptation, which adjusts to time-sensitive symptom contexts (e.g., "fever for 5 days"), and contextual precision, enabling differentiation of semantically and positionally complex diseases (e.g., chronic Arthritis vs. acute Dimorphic Hemorrhoids). However, limitations emerge in conditions with sparse training data, such as Cervical Spondylosis, and diseases with symptom overlap, like Typhoid vs. Urinary Tract Infection, pointing to opportunities for improvement. Integrating richer clinical data (e.g., laboratory results) or refining feature extraction techniques could further enhance performance. Overall, the DeBERTa with DCPG model validates its 99.8% accuracy, demonstrating its capacity to advance automated medical diagnostic systems with precision and adaptability.

### 4.3. Attention Map

In this subsection, we examine the explainability of the proposed method using an attention heatmap. For this purpose, the following sentence is utilized, and the corresponding attention heatmap is shown in Figure 5:

*"I'm currently experiencing a high fever, along with red spots and rashes covering my entire body. This has left me feeling extremely fatigued, with a significant loss of appetite, leading to weakness and a general sense of lethargy. I'm quite concerned about my condition."*

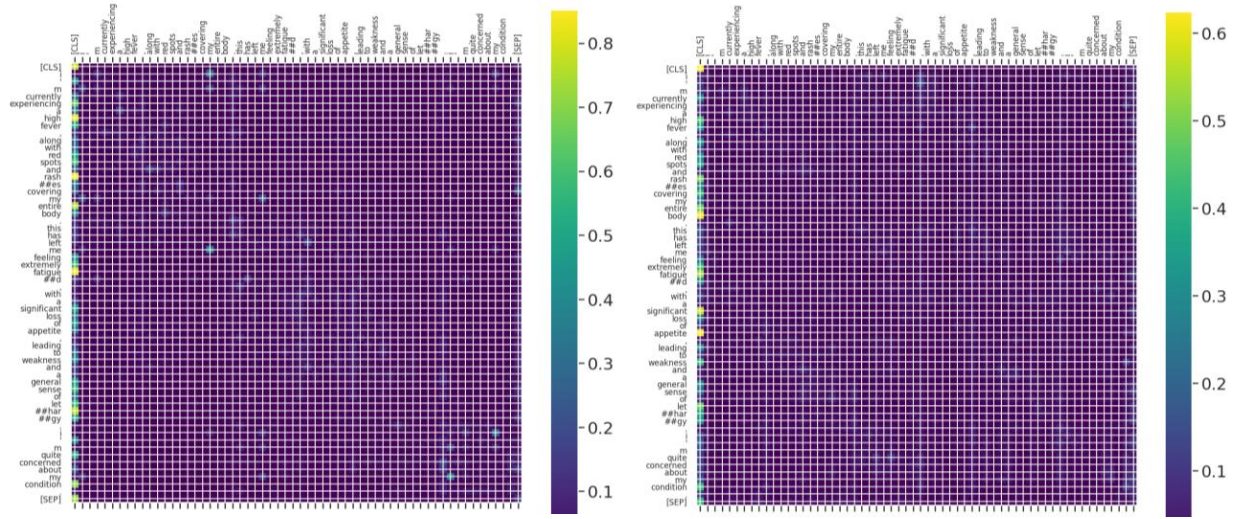

Figure 5: Attention heatmap using the proposed DeBERTa with DCPG.

The attention heatmap in Figure 5 demonstrates how the DeBERTa with DCPG model identifies and prioritizes key symptoms from the input text to classify the condition as chickenpox. High attention weights are assigned to phrases like "high fever" and "red spots and rashes covering my entire body," as these are hallmark features of chickenpox. The model captures the widespread and vesicular nature of the rash, along with systemic symptoms like "extreme fatigue" and "weakness," to differentiate chickenpox from other illnesses. By dynamically suppressing positional bias for generic terms (e.g., "loss of appetite") and amplifying attention for time-sensitive descriptors (e.g., "rash covering my entire body"), the DCPG mechanism ensures contextually precise symptom prioritization. This disentangled attention approach separates symptom content (e.g., "rash") from positional relevance (e.g., "entire body"), effectively capturing the progression and severity of chickenpox symptoms.

The model's performance aligns with clinical diagnostic guidelines for chickenpox, which include generalized vesicular rash and fever as key criteria. The heatmap further highlights its ability to distinguish chickenpox from other conditions like measles, allergy, or fungal infections by focusing on specific symptom patterns. For example, it rules out measles due to the absence of respiratory symptoms and eliminates allergy or fungal infections by ignoring localized descriptors like "swelling" or "scaly patches." This ability to adaptively emphasize critical symptoms, while maintaining explainability, enhances both diagnostic precision and clinical interpretability, making the DeBERTa-DCPG framework a valuable tool for automated medical diagnosis.

### 4.4. Quantitative Evaluation of the Classification Methods

This subsection provides a quantitative evaluation of the classification methods. Recall, Precision, and F1-score are crucial metrics for assessing classification models, particularly when dealing with imbalanced datasets. Recall, also known as sensitivity, measures the model's ability to correctly identify all relevant positive instances, emphasizing its effectiveness in capturing true positives. Precision, on the other hand, focuses on the accuracy of positive predictions, ensuring that the model minimizes false positives. The F1-score combines Recall and Precision into a single metric, offering a balanced evaluation of the model's performance in scenarios where both false positives and false negatives are important. Additionally, the Area Under the Curve (AUC) provides insight into the model's discrimination capability between positive and negative classes. A higher AUC value indicates better classification performance, with 1.0 representing a perfect classifier and 0.5 indicating random guessing. The following metrics represent the performance indices used to evaluate the models, with their corresponding results presented in Table 2.

$$FPR = \frac{FP}{FP + TN} \tag{19}$$

$$Recall = TPR = \frac{TP}{TP + FN} \tag{20}$$

$$ROC - AUC = \int_0^1 TPR(FPR) \, dFPR \tag{21}$$

$$Precision = \frac{TP}{TP + FP}. \tag{22}$$

$$F1 - score = 2 \frac{Precision \times Recall}{Precision + Recall}. \tag{23}$$

**Table 2.** Performance metrics of the proposed model compared to existing methods, including accuracy, precision, recall, and F1-score [11].

|  | Accuracy (%) | Recall (%) | Precision (%) | F1-score (%) | AUC-ROC |
|---|---|---|---|---|---|
| **BiLSTM + Hyperopt** [11] | 97.08 | 97.08 | 97.37 | 97.05 | 98.32 |
| **MCN-BERT + AdamW** [11] | 98.33 | 98.23 | 98.39 | 98.18 | 98.80 |
| **MCN-BERT + AdamP** [11] | 99.58 | 99.28 | 99.18 | 99.13 | 99.27 |
| **BT+BERT** | 97.85 | 97.52 | 96.95 | 97.23 | 98.45 |
| **BT + BERT + ABFNN** | 98.45 | 98.77 | 98.26 | 98.51 | 98.89 |
| **BT + DeBERTa + ABFNN** | 99.66 | 99.34 | 99.41 | 99.37 | 99.72 |
| **BT + DeBERTa - DCPG + ABFNN** | 99.78 | 99.72 | 99.79 | 99.75 | 99.88 |

The performance metrics highlight the superiority of the BT + DeBERTa - DCPG + ABFNN model in medical diagnosis tasks compared to other models. With an accuracy of 99.78%, recall of 99.72%, precision of 99.79%, F1-score of 99.75%, and AUC-ROC of 99.88%, it outperforms all the other approaches, including MCN-BERT + AdamP and BT + DeBERTa + ABFNN. The inclusion of the DCPG mechanism improves the model's ability to dynamically adjust positional relevance, leading to better classification of complex symptom patterns. This adaptability is particularly evident in the recall and F1-score, where the DCPG-based model excels, indicating a stronger balance between sensitivity and precision.

BiLSTM + Hyperopt: This model achieves moderate performance, with metrics around 97%, but its effectiveness is constrained by its limited ability to capture long-range dependencies in symptom descriptions. MCN-BERT + AdamW/AdamP: These models show improved performance, ranging from 98% to 99%, benefiting from BERT's powerful contextual embeddings. The AdamP optimizer slightly outperforms AdamW, likely due to its enhanced adaptive gradient clipping, which refines the optimization process for better convergence [11]. Compared to prior approaches such as MCN-BERT + AdamW/AdamP and BT + BERT + ABFNN, the BT + DeBERTa - DCPG + ABFNN model demonstrates incremental but significant gains in all metrics. While MCN-BERT models achieve high AUC-ROC values (up to 99.27%), they fall short in precision and recall compared to the DCPG-enhanced framework. Similarly, BT + BERT + ABFNN shows promising results with accuracy and recall around 98.45%, but it cannot match the contextual precision offered by DeBERTa and its DCPG mechanism. These results validate the effectiveness of the proposed model in capturing nuanced relationships between symptoms and diagnoses, making it a state-of-the-art solution for automated medical classification tasks.

## 5. Conclusion

This research introduces a groundbreaking NLP architecture that improved automated medical diagnosis. Our innovative approach combines Back-Translation techniques, an enhanced version of DeBERTa featuring DCPG, and an ABFNN. This synergistic integration results in a cutting-edge system that outperforms existing methods in medical text classification, setting a new benchmark in the field. The exceptional results, with an accuracy of 99.78%, recall of 99.72%, precision of 99.79%, and an F1-score of 99.75%, demonstrate the model's robust ability to accurately interpret and classify complex medical data. These metrics represent a substantial improvement over existing methods, including MCN-BERT and BiLSTM approaches, highlighting the effectiveness of our integrated framework. Key to this success is the synergistic combination of advanced NLP techniques. Back-Translation enhances data diversity, mitigating overfitting and improving generalization. DeBERTa with DCPG provides nuanced contextual embeddings, dynamically adjusting the influence of positional information based on semantic context. The ABFNN component effectively prioritizes diagnostically significant features, ensuring accurate predictions.

Future work will focus on expanding the model's dataset diversity, incorporating multilingual and rare disease cases for better generalization. Implementing real-time deployment in clinical settings and integrating explainable AI techniques will enhance trust and practical utility. Further research will aim to extend the model to handle multi-label classifications, include temporal data for disease progression, and explore lightweight architectures for more efficient deployment.

## Declarations

**Conflict of interest**: The authors have no relevant financial or nonfinancial interests to disclose.

**Funding:** The authors declare that no funds, grants, or other support were received during the preparation of this manuscript.

**Data availability:** The data that support the findings of this study are available as follows: https://www.kaggle.com/datasets/niyarrbarman/symptom2disease.